\documentclass[11pt,a4paper]{article}
\usepackage[hyperref]{naaclhlt2019}
\usepackage{times,graphicx, float}
\usepackage{latexsym, soul, xcolor, amsmath, amssymb}
\usepackage[inline]{enumitem}
\usepackage{url, float}
\usepackage{tikz}
\usepackage{array}
\usepackage{subcaption}
\aclfinalcopy 
\def\aclpaperid{541} 

\title{Neural Constituency Parsing of Speech Transcripts}

\author{Paria Jamshid Lou\\
  Macquarie University\\
  Sydney, Australia \\
 \\\And
  Yufei Wang \\
  Macquarie University\\
  Sydney, Australia\\ 
{\tt \hspace{-7cm} \{paria.jamshid-lou,yufei.wang\}@hdr.mq.edu.au} \\\AND
  Mark Johnson \\
  Macquarie University \\
  Sydney, Australia\\
  {\tt mark.johnson@mq.edu.au} \\}

\date{}

\begin{document}
\maketitle
\begin{abstract}
This paper studies the performance of a neural self-attentive parser on transcribed speech. Speech presents parsing challenges that do not appear in written text, such as the lack of punctuation and the presence of speech disfluencies (including filled pauses, repetitions, corrections, etc.).  Disfluencies are especially problematic for conventional syntactic parsers, which typically fail to find any EDITED disfluency nodes at all.  This motivated the development of special disfluency detection systems, and special mechanisms added to parsers specifically to handle disfluencies. However, we show here that neural parsers can find EDITED disfluency nodes, and the best neural parsers find them with an accuracy surpassing that of specialized disfluency detection systems, thus making these specialized mechanisms unnecessary. This paper also investigates a modified loss function that puts more weight on EDITED nodes. It also describes tree-transformations that simplify the disfluency detection task by providing alternative encodings of disfluencies and syntactic information\footnote{\url{https://github.com/pariajm/joint-disfluency-detector-and-parser}}.

\end{abstract}

\section{Introduction}
While a great deal of effort has been expended on parsing written text, parsing speech (either transcribed or ASR output) has received less attention. Parsing speech is important because speech is the easiest and most natural means of communication, it is increasingly used as an input modality in human-computer interactions. Speech presents parsing challenges that do not appear in written text, such as the lack of punctuation and sentence boundaries, speech recognition errors and the presence of speech disfluencies (including filled pauses, repetitions, corrections, etc.)~\citep{kahn:05}. Of the major challenges associated with transcribed speech, we focus here on speech disfluencies, which are frequent in spontaneous speech.

Disfluencies include filled pauses (``um'', ``uh''), parenthetical asides (''you know'', ''I mean''), interjections (``well'', ''like'') and partial words (``wou-'', ``oper-''). One type of disfluency which is especially problematic for conventional syntactic parsers are speech repairs. Following the analysis of~\citet{shri:94}, a speech repair consists of three main parts; the \textit{reparandum}, the \textit{interregnum} and the \textit{repair}. As illustrated in the following example, the reparandum \textit{we don't} is the part of the utterance that is replaced or repaired, the interregnum \textit{uh I mean} (which consists of a filled pause \textit{uh} and a discourse marker \textit{I mean}) is an optional part of the disfluency, and the repair \textit{a lot of states don't} replaces the reparandum. The fluent version is obtained by removing the reparandum and the interregnum.

\vspace{0.1cm}
\begin{equation}  \label{ex:1}
\centering 
\begin{array}{l}
\overbrace{\mbox{\it\strut \textcolor{blue}{We don't}}}^{\mbox{\scriptsize  \textcolor{blue}{reparandum}}}\strut\hspace{-0.5cm}~~~~~\overbrace{\mbox{\it \textcolor{red}{uh I mean}\strut}}^{\mbox{\scriptsize  \textcolor{red}{interregnum}}} 
\overbrace{\mbox{\it\strut a lot of states don't~}}^{\mbox{\scriptsize repair}}\\ \mbox{\it ~~~~~~~~~~~~have capital punishment.}
\end{array}
\end{equation}
\vspace{0.2cm}

\begin{figure} [h]
\centering 
\includegraphics[width=0.486\textwidth]{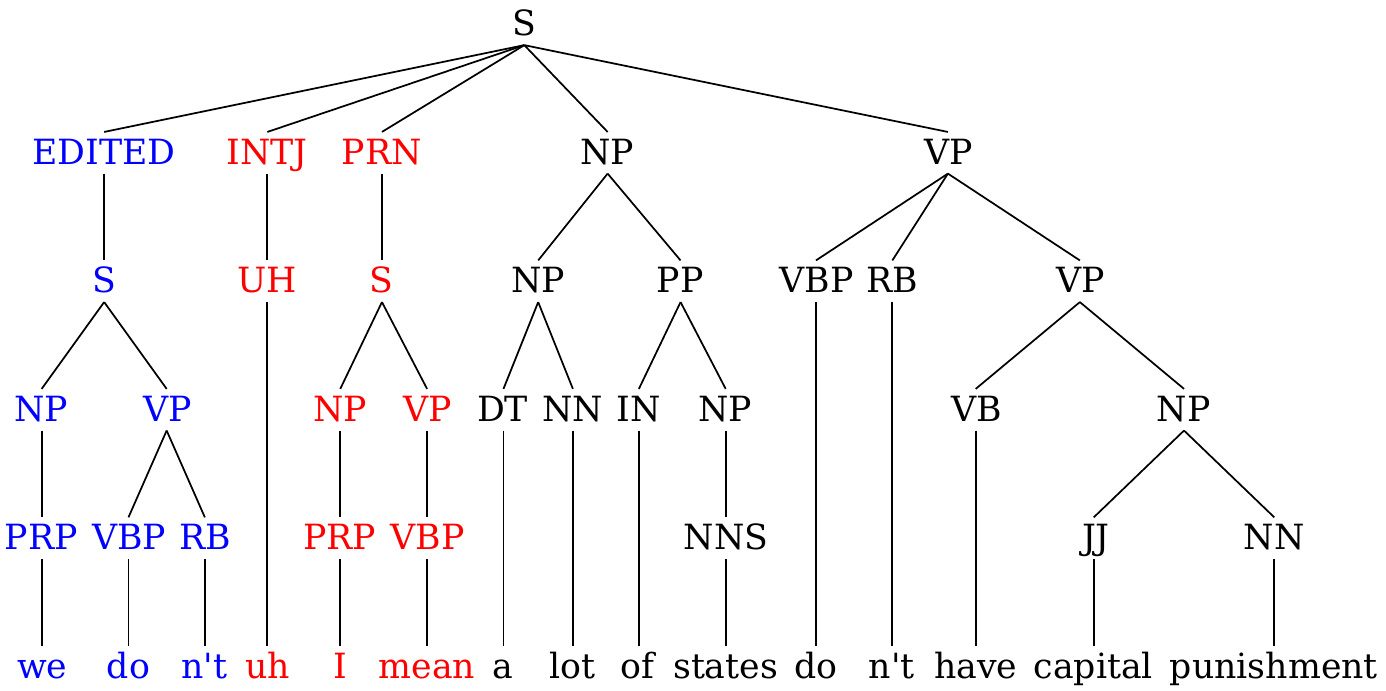}\caption{An example parse tree from the Switchboard corpus -- \emph{We don't uh I mean a lot of states don't have capital punishment,} where reparandum \emph{We don't}, filled pause \emph{uh} and discourse marker \emph{I mean} are dominated by EDITED, INTJ and PRN nodes.}
\label{fig:01}
\end{figure}

In the Switchboard treebank corpus~\citep{mit:99} the reparanda, filled pauses and discourse markers are dominated by EDITED, INTJ and PRN nodes, respectively (see Figure~\ref{fig:01}). Of these disfluency nodes, EDITED nodes pose a major problem for conventional syntactic parsers, as the parsers typically fail to find any EDITED nodes at all. Conventional parsers mainly capture tree-structured dependencies between words, while the relation between reparandum and repair is quite different: the repair is often a ``rough copy'' of the reparandum, using the same or very similar words in roughly the same order~\citep{char:01, john:04}. The ``rough copy'' dependencies are strong evidence of a disfluency, but conventional syntactic parsers cannot capture them. Moreover, the reparandum and the repair do not form conventional syntactic phrases, as illustrated in Figure~\ref{fig:01}, which is an additional difficulty when integrating disfluency detection with syntactic parsing. This motivated the development of special disfluency detection systems which find and remove disfluent words from the input prior to parsing~\citep{char:01, kahn:05, lease:06}, and special mechanisms added to parsers specifically to handle disfluencies~\citep{ras:13, hon:14, yoshi:16, trang:18}.   

\vspace{0.25cm}
\vspace{0.25cm}


In this paper, we investigate the performance of a neural self-attentive constituency parser on speech transcripts. We show that an ``off-the-shelf'' self-attentive parser, unlike conventional parsers, can detect disfluent words with a performance which is competitive to or better than specialized disfluency detection systems. In summary, the main contributions of this paper are:

\begin{itemize}
\item We show that the self-attentive constituency parser sets a new state-of-the-art for syntactic parsing of transcribed speech,
\item A neural constituency parser can detect EDITED words with an accuracy surpassing that of specialized disfluency detection models,
\item We demonstrate that syntactic information helps the neural syntactic parsing detect disfluent words more accurately,  
\item Replacing the constituent-based representation of disfluencies with a word-based representation of disfluencies improves the detection of disfluent words,
 
\item Modifying the training loss function to put more weight on EDITED nodes during training also improves disfluency detection.

\end{itemize}

\section{Related Work}
Speech recognition errors, unknown sentence boundaries and disfluencies are three major problems addressed by previous work on parsing speech. In this work, we focus on the problem of disfluency detection in parsing human-transcribed speech, where we assume that sentence boundaries are given and there are no word recognition errors. This section reviews approaches that add special mechanisms to parsers to handle disfluencies as well as specialized disfluency detection models.   

\subsection{Joint Parsing and Disfluency Detection}
Many speech parsers adopt a transition-based dependency approach to \begin{enumerate*}[label=(\roman*)] \item find the relationship between head words and words modifying the heads, and \item detect and remove disfluent words and their dependencies from the sentence. Transition-based parsers can be augmented with new parse actions to specifically handle disfluent words~\citep{ras:13, hon:14, yoshi:16, wu:15} \end{enumerate*}. A classifier is trained to choose between the standard and the augmented parse actions at each time step. Using pattern-match features in the classifier significantly improves disfluency detection~\citep{hon:14}. This reflects the fact that parsing based models use pattern-matching to capture the ``rough copy'' dependencies that are characteristic of speech disfluencies.

Speech parsing models usually use lexical features. One recent approach~\citep{trang:18} integrates lexical and prosodic cues in an encoder-decoder constituency parser. Prosodic cues result in very small performance gain in both parsing and disfluency detection. Augmenting the parser with a location-aware attention mechanism is specially useful for detecting disfluencies~\citep{trang:18}. 

In general, parsing models are poor at detecting disfluencies, mainly due to ``rough copy'' dependencies in disfluent sentences, which are difficult for conventional parsers to detect.

\subsection{Specialized Disfluency Detection Models}
Disfluency detection models often use a sequence tagging technique to assign a single label to each word of a sequence. Previous work shows that LSTMs and CNNs operating on words alone are poor at disfluency detection~\citep{zay:16, wang:16, jam:18}. The performance of state-of-the-art disfluency detection models depends heavily on hand-crafted pattern match features, which are specifically designed to find ``rough copies''. One recent paper~\citep{jam:18} augments a CNN model with a new kind of layer called an \emph{auto-correlational layer} to capture ``rough copy'' dependencies. The model compares the input vectors of words within a window to find identical or similar words. The addition of the auto-correlational layer to a ``vanilla'' CNN significantly improves the performance over the baseline CNN model. The results are competitive to models using complex hand-crafted features or external information sources, indicating that the auto-correlation model learns ``rough copies''. 

One recent paper~\citep{wang:18} introduces a semi-supervised approach to disfluency detection. Their self-attentive model is the current state-of-the-art result in disfluency detection. The common factor in~\citet{wang:18} and the approach presented here is the self-attentive transformer architecture, which suggests that this architecture is capable of detecting disfluencies with very high accuracy. The work we present goes beyond the work of~\citet{wang:18} in also studying the impact of jointly predicting syntactic structure and disfluencies (so it can be understood as a kind of multi-task learning). We also investigate the impact of different ways of representing disfluency information in the context of a syntactic parsing task. 

\section{Neural Constituency Parser}
We use the self-attentive constituency parser introduced by~\citet{kita:18} and train it on the Switchboard corpus of transcribed speech (we describe the training and evaluation conditions in more detail in Section~\ref{ex}). The self-attentive parser achieves state-of-the-art performance on WSJ data, which is why we selected it as the best ``off-the-shelf'' parsing model. The constituency parser uses a self-attentive transformer~\citep{vas:17} as an encoder and a chart-based parser~\citep{ster:17} as a decoder, as reviewed in the following sections. 

\subsection{Self-Attentive Encoder}
The encoder of a transformer is a stack of $n$ identical layers, each consists of two stacked sublayers: a multi-head attention mechanism, and a point-wise fully connected network. The inputs to the encoder first flow through a self-attention sublayer, which helps the encoder attends to several words in the sentence as it encodes a specific word. Because the model lacks recurrent layers, this sublayer is the only mechanism which propagates information between positions in the sentence. The self-attention maps the input to three vectors called query, key and value and defines an attention function as mapping a query and a set of key-value pairs to an output vector. The output is computed as a weighted sum of the values, where the weight assigned to each value is computed by a compatibility function of the query with the corresponding key. Each self-attention sublayer has several attention heads, where each head has its own query, key and value weight matrices. The multi-head attention allows the model to jointly attend to information from several different positions. The outputs of the self-attention layer are fed to a feed-forward neural network, which is applied to each position independently. For further detail, see~\citet{vas:17}. 

We believe that the self-attention mechanism is especially useful for detecting disfluencies in a sentence. In pilot experiments we found that similar LSTM-based parsers, such as the AllenNLP parser~\citep{gar:18}, were much worse at disfluency detection than the self-attentive parser.  As shown in Figure~\ref{fig:01}, the ``rough copy'' similarity between the repair and the reparandum is a strong indicator of disfluency. ``Rough copies'' involve same or very similar words in roughly same word order; for example, in the Switchboard training data, over $60\%$ of the words in the reparandum are exact copies of the words in the repair. Using the multi-head self-attention mechanism the model can presumably learn to focus on ``rough copies'' when detecting a reparandum. 

\subsection{Tree Score and Chart Parse Decoder}
A chart-based parser scores a tree as a sum of potentials on its labeled constituent spans as follows:
\begin{equation} \label{eq:02}
s(T)=\displaystyle\sum_{(i,j,l) \in T} s(i,j,l)
\end{equation}
where $s(i,j,l)$ is a score of a constituent  located between string positions $i$ and $j$ with label $l$. At test time, a modified CYK algorithm is used to find the highest scoring parse tree for a given sentence. 
\begin{eqnarray}
\hat{T} = \operatorname*{argmax}_{T} s(T)
\end{eqnarray}

Given the gold tagged tree $T^\star$, we train the model by minimizing a hinge loss:
\begin{equation} \label{eq:04}
max \bigg(0, \displaystyle{\max_{T\neq T^{\star}}}[s(T)+\bigtriangleup(T,T^{\star})]-s(T^{\star})\bigg)
\end{equation}
where $\bigtriangleup$ is the Hamming loss on labeled spans. For further detail, see~\citet{kita:18} and \citet{ster:17}.


\subsection{External Embedding and Edited Loss}
\citet{peters:18} have recently introduced a new approach for word representation called Embeddings from Language Models (ELMo) which has achieved state-of-the-art results in various NLP tasks. These embeddings are produced by a LSTM language model (LM) which inputs words and characters and generates a vector representation for each word of the sentence. The ELMo output is a concatenation of both the forward and backward LM hidden states. We found that using external ELMo embedding as the only lexical representation used by the model leads to the highest EDITED word f-score. Following~\citet{kita:18}, we use a trainable weight matrix to project the ELMo pretrained weights of 1024 dimension to a 512-dimensional content representation. We tried different combinations of input including predicted POS tags, character LSTM and word embeddings with ELMo, but the result was either worse or not significantly better than when using ELMo alone. 

The sole change we made to the self-attentive parser was to modify the loss function, so it puts more weight onto EDITED nodes. We show below that this improves the model's ability to recover EDITED nodes. We modify the tree scoring in~\ref{eq:02} as follows: 

\begin{equation}\label{eq:05}
s(T)=\displaystyle\sum_{(i,j,l) \in T} w_{l}~ s(i,j,l)
\end{equation} 
where $w_l$ depends on the label $l$. We only used two different values of $w_l$ here, one for EDITED nodes and one for all other node labels. We treat these as hyperparameters, and tune them to maximize EDITED nodes f-score (this is F(S$_\text{E}$) in Section~\ref{met} below).

\section{Experiments} \label{ex}
We evaluate the self-attentive parser on the Penn Treebank-3 Switchboard corpus~\cite{mit:99}. Following Charniak and Johnson~\shortcite{char:01}, we split the Switchboard corpus into training, dev and test sets as follows: training data consists of the sw[23]$\ast$.mrg files, dev data consists of the sw4[5-9]$\ast$.mrg files and test data consists of the sw4[0-1]$\ast$.mrg files. Except as explicitly noted below, we remove all partial words (words tagged XX and words ending in ``-'') and punctuation from data, as they are not available in realistic ASR applications~\citep{john:04}.

\subsection{Evaluation Metrics} \label{met}
We evaluate the self-attentive parser in terms of parsing accuracy and disfluency detection performance. We report \emph{precision} (P), \emph{recall} (R) and \emph{f-score} (F) for both \emph{constituent spans} (S) and \emph{word positions} (W), treating each word position as labeled by all the constituents that contain that word.  We also consider subsets of constituent spans and word positions; specifically:
\begin{enumerate*}[label=(\roman*)]
\item S$_\text{E}$, the set of constituent spans labeled EDITED,

\item W$_\text{E}$, the set of word positions dominated by one or more EDITED nodes, and

\item W$_\text{EIP}$, the set of word positions dominated by one or more EDITED, INTJ or PRN nodes.

\end{enumerate*}

\begin{figure}[h]
\centering
\begin{subfigure}{.247\textwidth}
  \centering
  \includegraphics[width=.927\linewidth]{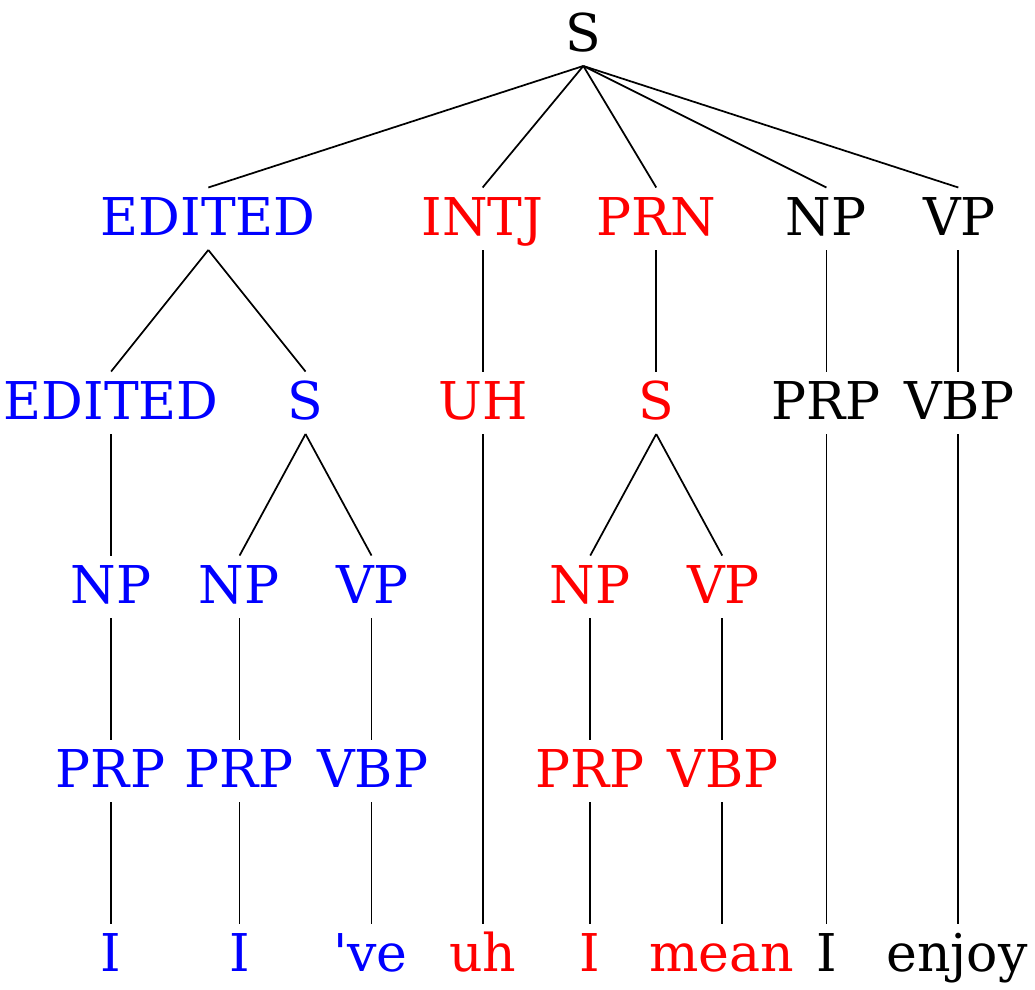}
  \caption{Gold tree}
  \label{fig:sub1}
\end{subfigure}%
\begin{subfigure}{.247\textwidth}
  \centering
  \includegraphics[width=.927\linewidth]{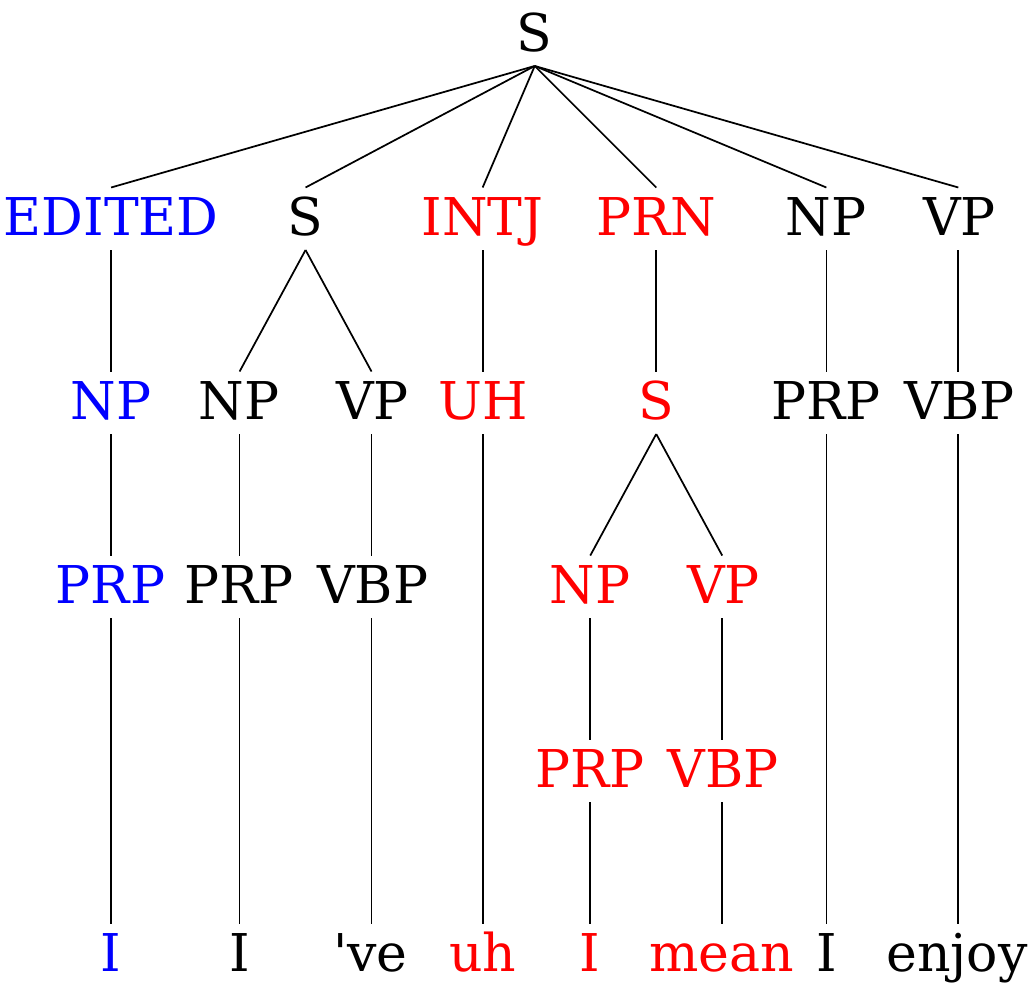}
  \caption{Predicted tree}
  \label{fig:sub2}
\end{subfigure}\vspace{0.4cm}

\emph{\small $_{0}$ I $_{1}$ I $_{2}$ 've $_{3}$ uh $_{4}$  I $_{5}$ mean $_{6}$ I $_{7}$ enjoy $_{8}$}
\caption{An example gold and predicted parse tree.}
\label{fig:test}
\end{figure}

We demonstrate the evaluation metrics with an example here. Consider the gold and predicted parse trees illustrated in Figure~\ref{fig:test}. The constituency trees are viewed as a set of labeled spans over the words of the sentence, where constituent spans are pairs of string positions. As explained earlier, we ignore punctuation and partial words when calculating evaluation scores. To calculate fscore for a span, i.e., F(S), the gold, predicted and correct labeled spans are counted. In this case, the number of predicted, gold and correctly predicted spans is $13$, $14$ and $12$.

Since a parse tree with EDITED nodes identifies certain words as EDITED, we can evaluate how accurately a parser classifies words as EDITED (i.e. F(W$_\text{E}$)). Continuing with the example in Figure~\ref{fig:test}, the number of predicted, gold and correctly predicted EDITED words is $1$, $3$ and $1$.

Similarly, we can also measure how well the parser can identify all disfluency words, i.e., the words dominated by EDITED, INTJ or PRN nodes. 
Continuing with the example in Figure~\ref{fig:test}, the number of predicted, gold and correctly predicted EDITED, INTJ and PRN words is $4$, $6$, $4$.

\subsection{Model Training} \label{con}
We use randomized search~\citep{berg:12} to tune the optimization and architecture parameters of the model on the dev set. We optimize the model for its performance on parsing EDITED nodes F(S$_\text{E}$). The hyperparameters include dimensionality of the model, learning rate, edited loss weight, dropout, number of layers and heads as shown in Table~\ref{tab:0ff}. All other hyperparameters not mentioned here are the same as in~\citet{kita:18}.   

\begin{table}[h]
\begin{center}
\begin{tabular}{|l|c|}
\hline \bf \small Configuration & \bf  \small Parser  \\ \hline
\small hidden label dim & \small$340$ \\ \hline
\small model dim & \small$2048$ \\ \hline
\small non-EDITED label weight & \small$0.7$ \\ \hline
\small EDITED label weight & \small$2$ \\ \hline
\small learning rate & \small$0.0006$ \\ \hline
\small learning rate warmup steps & \small$110$ \\ \hline
\small step decay factor & \small$0.52$ \\ \hline
\small num heads & \small$7$ \\ \hline
\small num layers & \small$4$ \\ \hline
\small attention dropout & \small$0.27$   \\ \hline
\small relu dropout & \small$0.09$ \\ \hline
\small residual dropout & \small$0.26$ \\ \hline
\small elmo dropout & \small$0.57$ \\ \hline
\small tag embedding dropout & \small$0.35$ \\ \hline
\small word embedding dropout & \small$0.2$ \\ \hline
\end{tabular}
\end{center}
\caption{Hyperparameter setting for the self-attentive constituency parser. }\label{tab:0ff} 
\end{table}

\subsection{Edited Loss}
Our best dev model (see Table~\ref{tab:0ff}) uses an edited loss that puts more weight on EDITED nodes and less weight on non-EDITED nodes. To explore the effect of edited loss, we retrained the best model with an equally weighted loss. The results in Table~\ref{tab:13} indicate that differential weighting improves parsing EDITED nodes as well as EDITED word detection. It also rebalances the precision vs. recall trade-off and slightly increases overall parsing accuracy F($\text{S}$).
 
\begin{table}[h]
\begin{center}
\begin{tabular}{|l|c|c|} \hline
 & \small \bf{equal weight} & \small \bf{different weight} \\ \hline
\bf \small P($\text{S}_\text{E}$) & \small$83.0$ & \small$83.3$ \\ \hline
\bf \small R($\text{S}_\text{E}$)  & \small$91.6$ & \small$91.4$ \\ \hline
\bf \small F($\text{S}_\text{E}$) & \small$87.1$ &  \small$87.2$ \\ \hline
\bf \small F($\text{S}$) & \small$92.8$  &  \small$93$   \\ \hline
\bf \small F($\text{W}_\text{E}$) & \small$86.9$ & \small$87.5$ \\ \hline 
\end{tabular}
\end{center}
\caption{\label{tab:13} Parsing precision P($\text{S}_\text{E}$), recall {R($\text{S}_\text{E}$)} and f-score {F($\text{S}_\text{E}$)} of EDITED nodes, parsing f-score {F($\text{S}$)} and EDITED word f-score {F($\text{W}_\text{E}$)} on the Switchboard dev set for the equally and differentially weighted loss.}
\end{table}

\subsection{Modifying the Training Data}
We investigate the effect of modifying the training data on the performance of the parser.

\subsubsection{Simplified Tree Structures}
We use different tree-transformations to explore the effect of different amounts of and encodings of disfluencies and syntactic information on the performance of the model.

\begin{itemize}

\item {\bf Baseline:} Parse trees as they appear in the Switchboard corpus. A sample is shown in Figure~\ref{fig:02}.
\begin{figure} [h!]
\centering 
\includegraphics[width=0.35\textwidth]{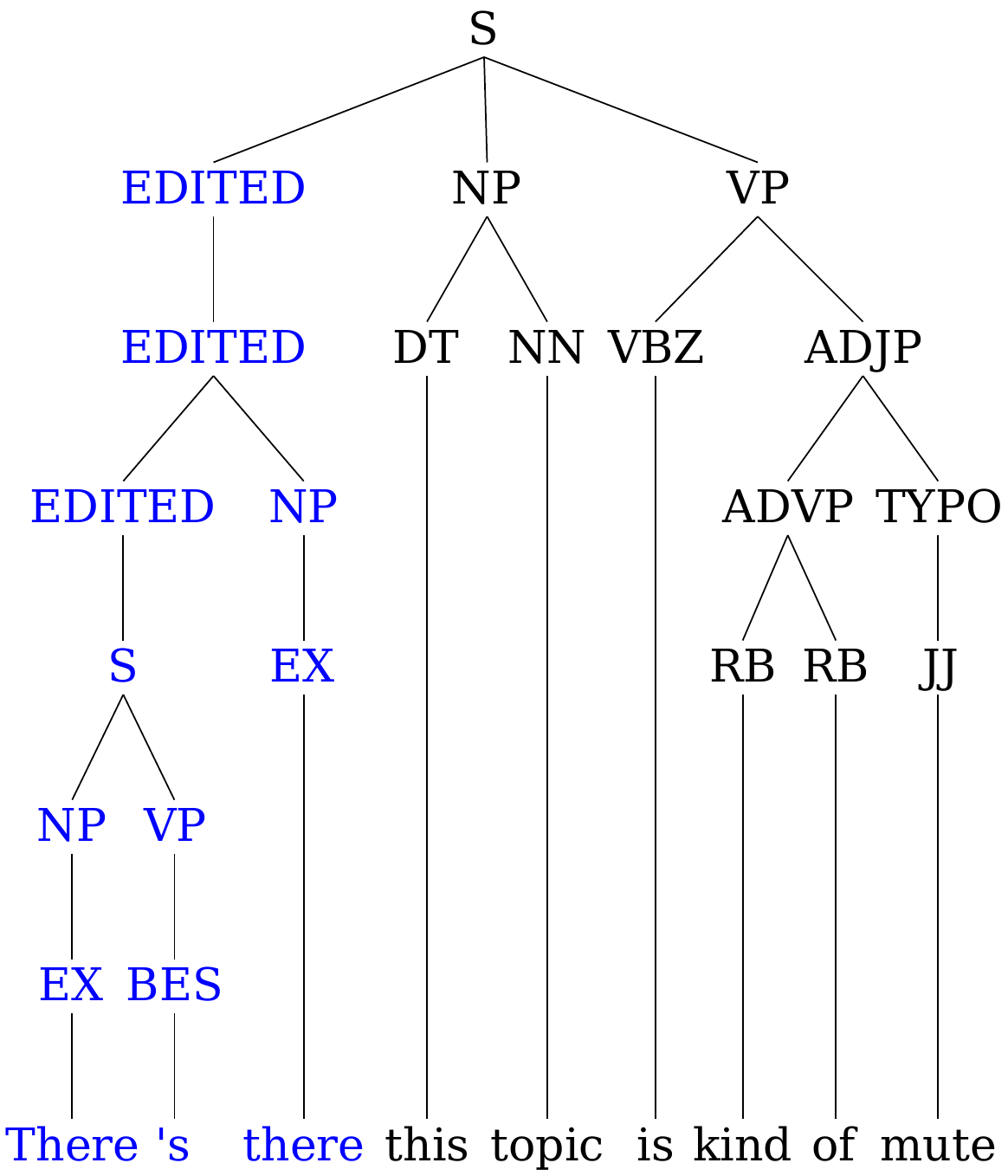} \caption{Baseline parse trees as they appear in the Switchboard corpus} \label{fig:02}
\end{figure}

\item {\bf Transformation PosDisfl:} Pushing disfluency nodes (i.e. EDITED, INTJ and PRN) down to POS tags, as shown in Figure~\ref{fig:03}.
\begin{figure} [h!]
\centering 
\includegraphics[width=0.4\textwidth]{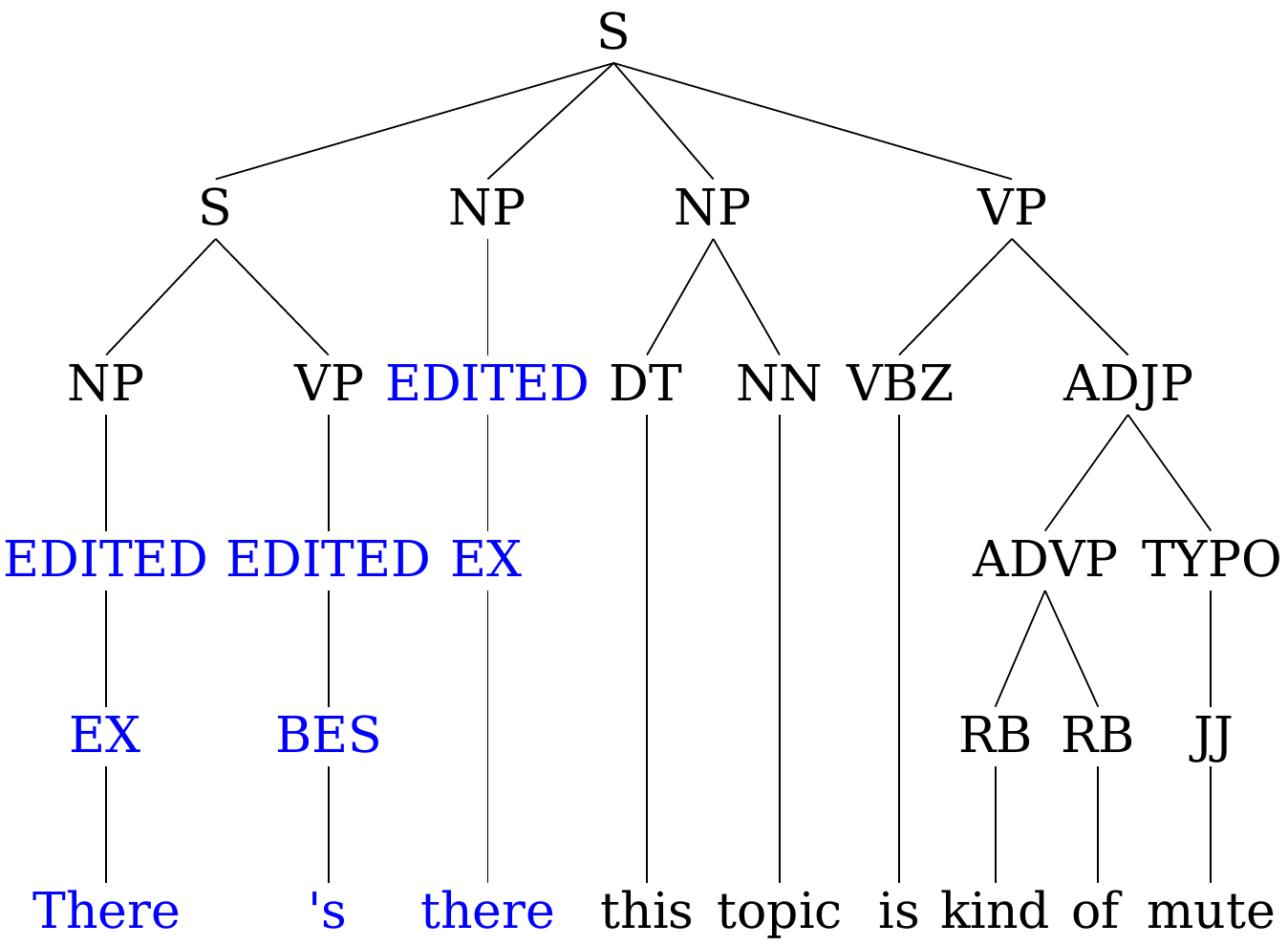} \caption{Transformation PosDisfl, where disfluency nodes are pushed down to POS tags.} \label{fig:03}
\end{figure}

\item {\bf Transformation NoSyntax:} Deleting all non-disfluency nodes, as shown in Figure~\ref{fig:04}.
\begin{figure} [h!]
\centering 
\includegraphics[width=0.37\textwidth]{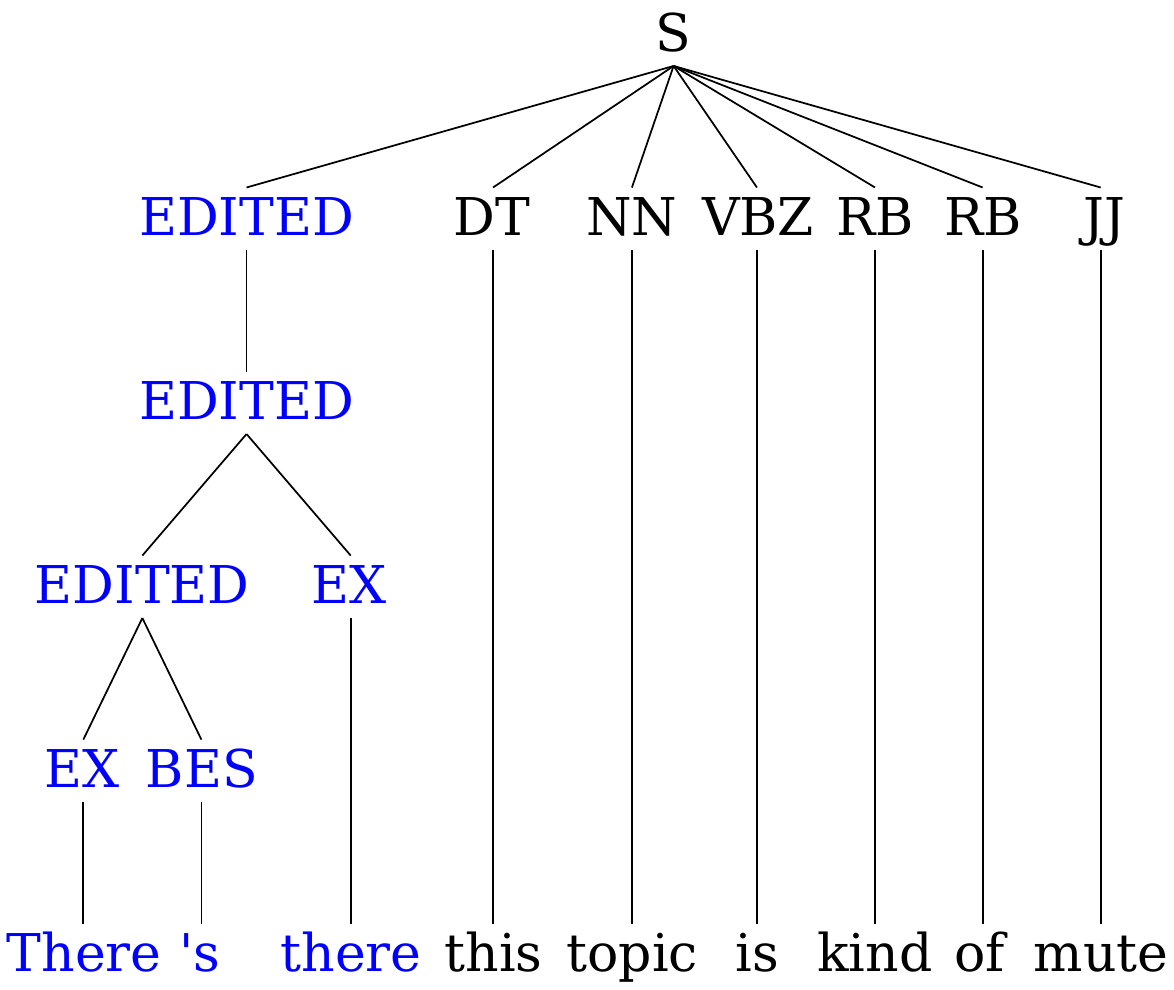} \caption{Transformation NoSyntax, where all non-disfluency nodes are deleted.} \label{fig:04}
\end{figure}

\item {\bf Transformation PosDisfl+NoSyntax:} Pushing disfluency nodes down to POS tags and deleting all non-disfluency nodes, as shown in Figure~\ref{fig:05}.
\begin{figure} [h!]
\centering 
\includegraphics[width=0.45\textwidth]{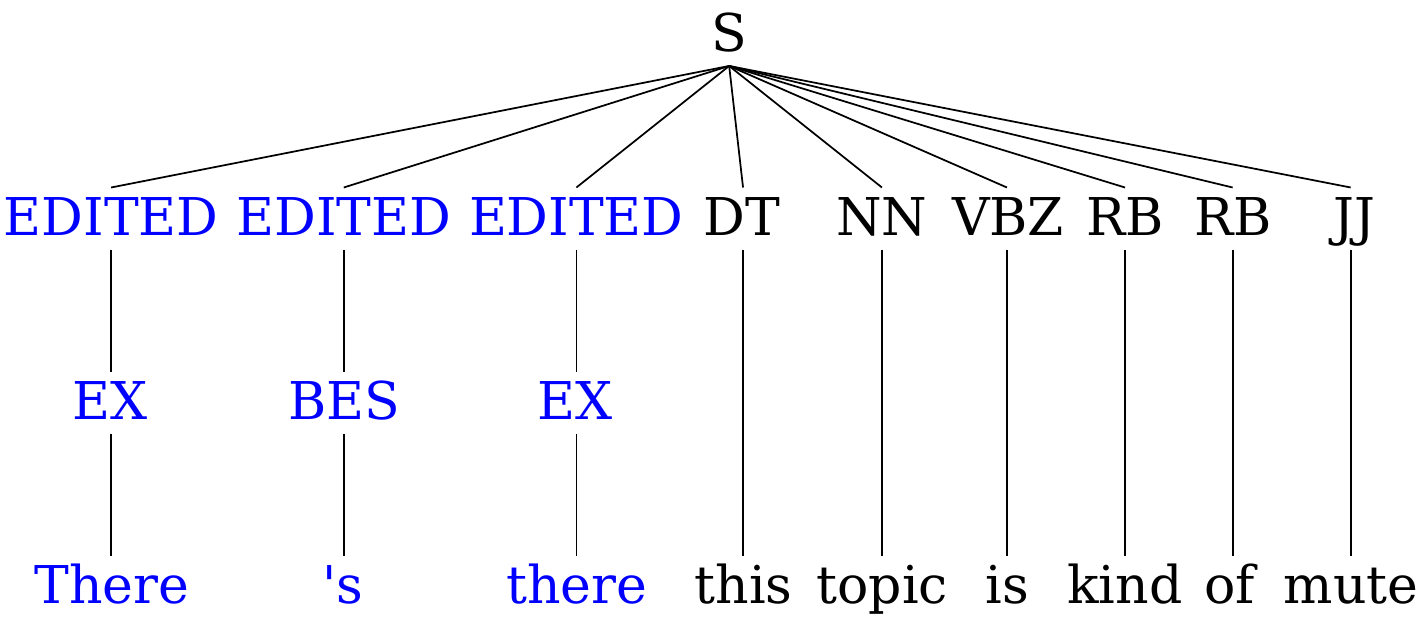} \caption{Transformation PosDisfl+NoSyntax, where disfluency nodes are pushed down to POS tags and all non-disfluency nodes are deleted.} \label{fig:05}
\end{figure}

\item {\bf Transformation TopDisfl:} Deleting all disfluency nodes but the top ones, as shown in Figure~\ref{fig:06}.
\begin{figure} [h!]
\centering 
\includegraphics[width=0.35\textwidth]{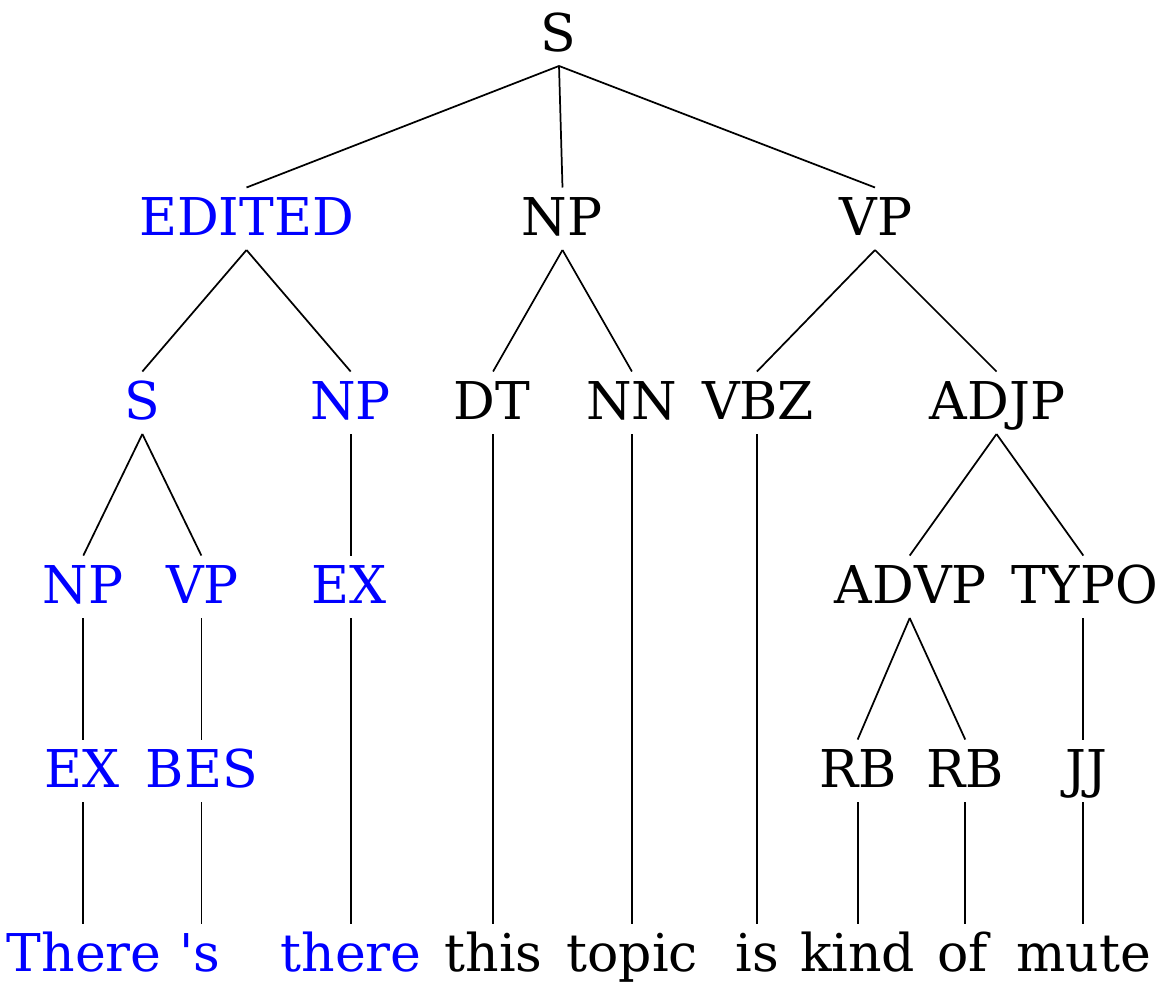} \caption{Transformation TopDisfl, where all disfluency nodes but the top ones are deleted.} \label{fig:06}
\end{figure}

\item {\bf Transformation TopDisfl+NoSyntax:} Deleting all nodes but the top-most disfluency nodes, as shown in Figure~\ref{fig:07}.
\begin{figure} [h!]
\centering 
\includegraphics[width=0.4\textwidth]{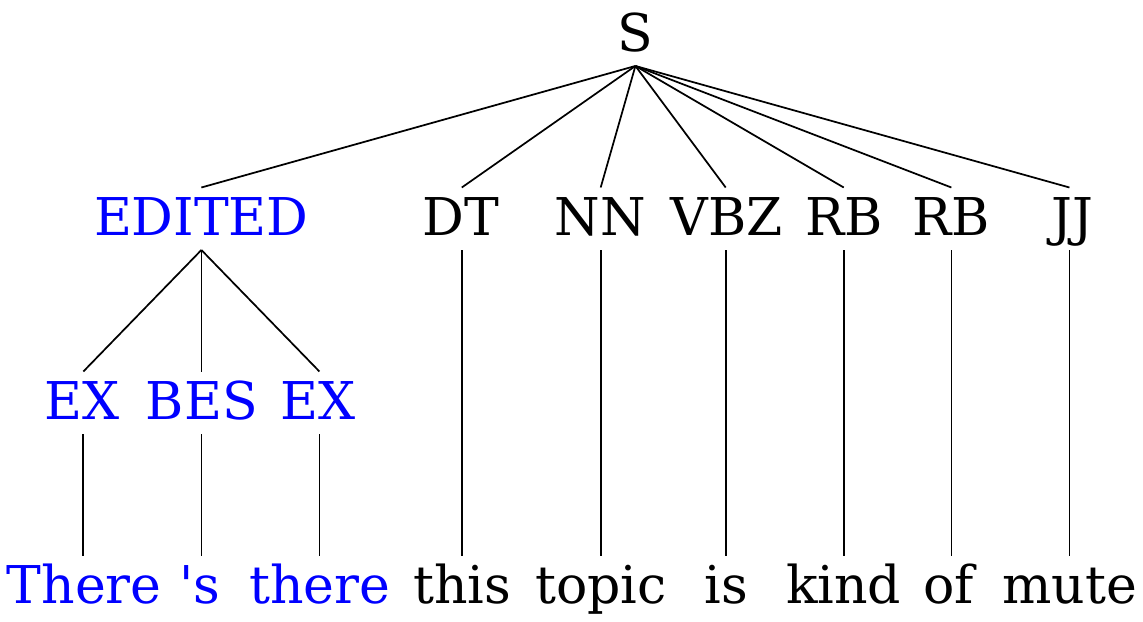}
\caption{Transformation TopDisfl+NoSyntax, where all nodes but the top-most disfluency nodes are deleted.} \label{fig:07}

\end{figure}
\end{itemize}

We report the performance of the self-attentive parser in terms of EDITED word f-score and disfluency word f-score in Table~\ref{tab:05}. Since the transformations change the tree shapes, it is not meaningful to compare their parsing f-scores. As illustrated in Table~\ref{tab:05}, pushing disfluency nodes down to POS tags (i.e. \emph{Transformation PosDisfl}) increases precision about $2\%$, resulting in $1\%$ improvement in word f-score F(W$_\text{E}$). It also improves F(W$_\text{EIP}$) by $0.4\%$. In general, the model can take advantage of the simplified encoding of disfluency nodes (see \emph{Transformations PosDisfl} and \emph{TopDisfl}). Moreover, deleting all but the top-most disfluency nodes as in \emph{Transformation TopDisfl+NoSyntax} significantly drops precision (about $20\%$), resulting in more than $13\%$ decrease in EDITED word f-score. It also hurts detecting all types of disfluency (more than $7\%$ decrease in F(W$_\text{EIP}$)). In general, removing syntactic structure dramatically degrades the performance of the model in terms of F(W$_\text{E}$) and F(W$_\text{EIP}$), as shown in \emph{Transformations NoSyntax}, \emph{PosDisfl+NoSyntax} and \emph{TopDisfl+NoSyntax}. This indicates that syntactic information is important for detecting disfluencies.  
 
\begin{table}[h]
\begin{center}
\begin{tabular}{|>{\raggedright}p{2.36cm}|>{\centering}p{0.7cm}|>{\centering}p{0.76cm}|>{\centering}p{0.72cm}|>{\centering\arraybackslash}p{0.9cm}|}
\hline \small \bf Setting & \bf \small P(W$_\text{E}$) & \bf \small R(W$_\text{E}$) & \bf \small F(W$_\text{E}$) & \bf \small F(W$_\text{EIP}$)  
\\ \hline 
\small Baseline & \small$81.6$ & \small$94.2$ & \small$87.5$ &  \small$94.0$\\ \hline
\small PosDisfl & \small$83.7$ & \small$94.2$  &   \small$\bf88.7$  & \small$\bf94.4$  \\ \hline
\small NoSyntax & \small$73.0$ & \small$95.0$ &  \small $82.5$  & \small$92.3$ \\ \hline
\small PosDisfl+NoSyntax & \small$73.4$  &\small $93.4$ &  \small$82.2$  & \small$91.6$ \\ \hline
\small TopDisfl & \small$81.9$ & \small$94.3$ &  \small$87.7$  & \small$93.8$ \\ \hline
\small TopDisfl+NoSyntax & \small$61.3$ & \small$93.1$ & \small $74.0$ & \small$86.7$ \\ \hline
\end{tabular}
\end{center}
\caption{\label{tab:05} EDITED word precision {P(W$_\text{E}$)}, recall {R(W$_\text{E}$)} and f-score {F(W$_\text{E}$)} as well as EDITED, INTJ and PRN word f-score {F(W$_\text{EIP}$)} on the Switchboard dev set for different encodings of disfluency nodes in data. The best f-scores are shown in bold.}
\end{table}

\subsubsection{Punctuation and Partial words}
As mentioned before, speech recognition models generally do not produce punctuation and partial words in their outputs. Thus, prior work has  removed them from the data to make the evaluation more realistic. However, it is interesting to see what information partial words and punctuation convey about syntactic structure in general and disfluencies in particular, so we did an experiment to investigate the effect of including these in the training and test data. 
We use the best hyperparameter configuration on the Switchboard dev set and retrain the model on two versions of the data: \begin{enumerate*}[label=(\roman*)]\item with partial words and \item with punctuation and partial words.  \end{enumerate*} As shown in Table~\ref{tab:04}, keeping punctuation and partial words in the training data increases EDITED word f-score by about $4\%$, indicating that punctuation and partial words greatly help disfluency detection. Punctuation leads to more gain in disfluency detection than partial words. Punctuation also improves the word f-score for all types of disfluencies by more than $1\%$. 

\begin{table}[h]
\begin{center}
\begin{tabular}{|p{3.5cm}|>{\centering}p{0.8cm}|>{\centering\arraybackslash}p{1cm}|}
\hline \bf \small Setting & \bf  \small F($\text{W}_\text{E}$) & \bf  \small F($\text{W}_\text{EIP}$) \\ \hline 
\small without punctuation \& partial words  & \small$88.7$ &  \small$94.4$ \\ \hline
\small with partial words  & \small$89.7$ & \small$94.4$ \\ \hline
\small with punctuations \& partial words  & \small$92.2$ & \small$95.5$  \\ \hline
\end{tabular}
\end{center}
\caption{\label{tab:04} EDITED word {F($\text{W}_\text{E}$)} and EDITED, INTJ and PRN word f-score {F($\text{W}_\text{EIP}$)} on the Switchboard dev set for three versions of the training data.}
\end{table}

\section{Results}
We selected our best model based on the dev set results (including differentially weighted loss) and compared the results achieved for the \emph{Tree Transformation PosDisfl} and \emph{No Tree Transformation} on the test set with previous work. Although most previous work has used the Switchboard corpus, it is sometimes difficult to compare systems directly due to different scoring metrics and differences in experimental setup, such as the use of partial words, punctuation, prosodic cues and so on. Since some studies report their results using partial words and/or punctuation, we divide prior work according to the setting they used and report the results of the self-attentive parser on the test data for each setting.

Table~\ref{tab:01} shows the test performance of the self-attentive constituency parser against previous parsing models of speech transcripts. The self-attentive parser outperforms all previous models in parsing accuracy. It has also better performance than~\citet{kahn:05} and \citet{trang:18}, who used acoustic/prosodic cues from speech waveform as well as the words in the transcript. 

\begin{table}[h!]
\begin{center}
\begin{tabular}{|>{\raggedright}p{4.2cm}|c|}
\hline \bf \footnotesize Parsing Model  &  \footnotesize\bf F(S)  \\ \hline

\small{without partial words} &  \\ 
~~~\small{self-attentive parser (PDT) }&  \small$92.4$ \\
~~~\bf\small{self-attentive parser (NT) }&  \small$\bf92.7$ \\ \hline
\small{partial words:} &  \\ 
~~~\small\citet{hale:06} &  \small$71.1$ \\ 
~~~\small\citet{kahn:05} &  \small$86.6$ \\ 
~~~\small\citet{trang:18}&    \small$88.5$ \\
~~~\small{self-attentive parser (NT)}& \small$92.3$ \\
~~~\bf\small{self-attentive parser (PDT)}& \small$\bf92.6$ \\ \hline
\end{tabular}
\end{center}
\caption{\label{tab:01} Parse f-score F(S) for all constituent spans on the Switchboard test set with and without partial words. NT = No Transformation and PDT = PosDisfl Transformation.}
\end{table}

We also compare the performance of the self-attentive parser with state-of-the-art disfluency detection methods in terms of EDITED word f-score. As shown in Table~\ref{tab:02}, the self-attentive parser (with \emph{PosDisfl Transformation}) achieves a new state-of-the-art for detecting EDITED words. Its performance is competitive with specialized disfluency detection models that directly optimize for disfluency detection. Using partial words increases edited word f-score for \emph{No Transformation} mode by $0.1\%$ and for \emph{PosDisfl Transformation} mode by $0.6\%$, which is not surprising as the presence of partial words is strongly correlated with the presence of a disfluency.

It is interesting to compare the self-attentive parser with the ACNN model presented in~\citet{jam:18}. They introduce a new ACNN layer which is able to learn the ``rough copy'' dependencies between words, for which previous models heavily relied on hand-crafted pattern-matching features. ``Rough copies'' are a strong indicator of disfluencies that can help the model detect reparanda (i.e. EDITED nodes). That the self-attentive parser is better than the ACNN model~\citep{jam:18} in detecting disfluencies may indicate that the self-attention mechanism can learn ``rough copy'' dependencies. 

\begin{table}[h!]
\begin{center}
\begin{tabular} {|>{\raggedright}p{5cm}|c|} \hline \bf \footnotesize Model  & \footnotesize\bf F(W$_\text{E}$) \\ \hline 
\small{without partial words:} & \\  
~~~\small\citet{hon:14} &  \small$84.1$  \\ 
~~~\small\citet{jam:18}~\small{$\bullet$} & \small$84.5$  \\ 
~~~\small\citet{wu:15} & \small$85.1$ \\ 
~~~\small\citet{fer:15}~\small{$\bullet$} & \small$85.4$  \\ 
~~~\small\citet{wang:16}~\small{$\bullet$} &\small$86.7$ \\  
~~~\small\citet{jam:17}~\small{$\bullet$} & \small $86.8$ \\  
~~~\small{self-attentive parser (NT)}&  \small$86.9$\\ 
~~~\small\citet{wang:17} ~\small{$\bullet$} & \small $87.5$ \\  
~~~\bf\small{self-attentive parser (PDT)}&  \small$\bf88.1$\\ 

\hline
\small{partial words:} & \\  
~~~\small\citet{hale:06}  & \small$41.7$ \\ 
~~~\small\citet{trang:18}  & \small$77.5$ \\ 
~~~\small\citet{kahn:05} & \small$78.2$ \\
~~~\small\citet{ras:13} & \small$81.4$  \\  
~~~\small\citet{zay:16}~\small{$\bullet$}& \small$85.9$ \\  
~~~\small{self-attentive parser (NT)} & \small$87.0$\\ 
~~~\bf\small{self-attentive parser (PDT)} & \small$\bf88.7$\\ \hline

\end{tabular}
\end{center}
\caption{\label{tab:02} Edited word f-score {{F(W$_\text{E}$)}} on the Switchboard test set with and without partial words. {~\small$\bullet$} Specialized disfluency detection models. NT = No Transformation and PDT = PosDisfl Transformation. }
\end{table}

We also compare the performance of the self-attentive parser with Wang et al.'s~\citeyearpar{wang:18} self-attentive disfluency detection model in terms of disfluency (i.e. EDITED, INTJ and PRN) word f-score. As shown in Table~\ref{tab:11}, the self-attentive parser outperforms this state-of-the-art specialized self-attentive disfluency detection model. 

\begin{table}[h!]
\begin{center}
\begin{tabular}{|p{4.5cm}|c|}
\hline \bf \footnotesize Self-attentive Model  & \small \bf F(W$_\text{EIP}$)  \\ \hline
\small{punctuation \& partial words}  & \\ 
~~~\small\citet{wang:18}& \small$91.1$ \\ 
~~~\small{self-attentive parser (NT)}&\small$93.7$ \\ 
~~~\bf\small{self-attentive parser (PDT)}&\small$\bf94.0$ \\ \hline
\end{tabular}
\end{center}
\caption{\label{tab:11} EDITED, INTJ and PRN word f-score {F(W$_\text{EIP}$) } on the Switchboard test set with punctuation and partial words. NT = No Transformation and PDT = PosDisfl Transformation.}
\end{table}





\section{Conclusion and Future Work}
This paper shows that using an ``off-the-shelf'' constituency parser achieves a new state-of-the-art in parsing transcribed speech. The self-attentive parser is effective in detecting disfluent words as it outperforms specialized disfluency detection models, suggesting that it is feasible to use standard neural architectures to perform disfluency detection as part of some other task, rather than requiring a separate disfluency detection pre-processing step. We also show that removing syntactic information hurts word f-score. That is, performing syntactic parsing and disfluency detection as a multi-task training objective yields higher disfluency detection accuracy than performing disfluency detection in isolation. Modifying encoding by indicating disfluencies at the word level leads to further improvements in disfluency detection. 

In future work we hope to integrate syntactic parsing more closely with automatic speech recognition. A first step is to develop parsing models that parse ASR output, rather than speech transcripts. It may also be possible to more directly integrate an attention-based syntactic parser with a speech recogniser, perhaps trained in an end-to-end fashion.

\section*{Acknowledgments}

This research was supported by a Google award through the Natural Language Understanding Focused Program, CRP 8201800363 from Data61/CSIRO, and under the Australian Research Council’s Discovery Projects funding scheme (project number DP160102156). We also thank the anonymous reviewers for their valuable comments that helped to improve the paper.

\bibliography{naaclhlt2019}
\bibliographystyle{acl_natbib}


\end{document}